\documentclass[letterpaper]{article} 
\usepackage{aaai24}  
\usepackage{times}  
\usepackage{helvet}  
\usepackage{courier}  
\usepackage[hyphens]{url}  
\usepackage{graphicx} 
\usepackage{natbib}  
\usepackage{caption} 
\usepackage{dsfont}
\usepackage{amsmath}
\usepackage{amsfonts}
\usepackage{algorithm}
\usepackage{algorithmic}

\usepackage{makecell}
\usepackage{multirow}

\usepackage{newfloat}
\usepackage{listings}
\DeclareCaptionStyle{ruled}{labelfont=normalfont,labelsep=colon,strut=off} 
\floatstyle{ruled}
\newfloat{listing}{tb}{lst}{}
\floatname{listing}{Listing}
\title{STAIR: Spatial-Temporal Reasoning with Auditable Intermediate Results for Video Question Answering}

\author{
    Yueqian Wang\textsuperscript{\rm 1},
    Yuxuan Wang\textsuperscript{\rm 2,3},
    Kai Chen\textsuperscript{\rm 4},
    Dongyan Zhao\textsuperscript{\rm 1,3}\thanks{Corresponding Author}
}

\affiliations{

    \textsuperscript{\rm 1}Wangxuan Institute of Computer Technology, Peking University\\
    \textsuperscript{\rm 2}Beijing Institute for General Artificial Intelligence \\
    \textsuperscript{\rm 3}National Key Laboratory of General Artificial Intelligence\\
    \textsuperscript{\rm 4}School of Economics, Peking University\\
    wangyueqian@pku.edu.cn, wangyuxuan1@bigai.ai, chen.kai@pku.edu.cn, zhaodongyan@pku.edu.cn

}

\begin{document}

\maketitle

\begin{abstract}
Recently we have witnessed the rapid development of video question answering models. However, most models can only handle simple videos in terms of temporal reasoning, and their performance tends to drop when answering temporal-reasoning questions on long and informative videos. 
To tackle this problem we propose \textbf{STAIR}, a \textbf{S}patial-\textbf{T}emporal Reasoning model with \textbf{A}uditable \textbf{I}ntermediate \textbf{R}esults for video question answering. STAIR is a neural module network, which contains a program generator to decompose a given question into a hierarchical combination of several sub-tasks, and a set of lightweight neural modules to complete each of these sub-tasks.
Though neural module networks are already widely studied on image-text tasks, applying them to videos is a non-trivial task, as reasoning on videos requires different abilities. In this paper, we define a set of basic video-text sub-tasks for video question answering and design a set of lightweight modules to complete them.
Different from most prior works, modules of STAIR return intermediate outputs specific to their intentions instead of always returning attention maps, which makes it easier to interpret and collaborate with pre-trained models. We also introduce intermediate supervision to make these intermediate outputs more accurate. We conduct extensive experiments on several video question answering datasets under various settings to show STAIR's performance, explainability, compatibility with pre-trained models, and applicability when program annotations are not available. Code: https://github.com/yellow-binary-tree/STAIR
\end{abstract}

\section{Introduction}
Video question answering (video QA) is a challenging task that lies between the field of Natural Language Processing and Computer Vision, which requires a joint understanding of text and video to give correct answers.
However, most approaches, including some recently proposed video-text large pre-trained models, only treat videos as animated images. They use black-box deep neural networks to learn mappings directly from inputs to outputs on factual questions like ``Who is driving a car?'', ignoring the biggest difference between videos and images: the existence of temporal information. As a result, their performance tends to drop when understanding long and informative videos and answering complicated temporal-reasoning questions, such as determining the order of two events, or identifying events in a given time period of the video, where small differences in temporal expressions can lead to different results.

In comparison, in image question answering, many neural-symbolic methods have been proposed to tackle with complicated spatial-reasoning problems. Neural Symbolic VQA \cite{yi2018neuralSymbolic} aims to parse a symbolic scene representation out of an image, and converts the question to a program that executes on the symbolic scene representation. Neural Symbolic Concept Learners \cite{Mao2019TheNC} also convert images to symbolic representations, but by learning vector representations for every visual concept.
However, though these neural symbolic methods can achieve very good results on synthetic images like CLEVR \cite{Johnson2016CLEVRAD} and Minecraft \cite{wu2017neural,yi2018neuralSymbolic}, they can not perform well on real-world images. One promising neural-symbolic approach is Neural Module Networks (NMNs) \cite{Andreas2015NeuralMN}. It first converts the question to a program composed of several functions using a program generator, and then executes the program by implementing each function with a neural network, which is also known as a ``module''. With the introduction of neural networks at execution, it works better on real-word image question answering like VQA \cite{Agrawal2015VQAVQ}, and can also provide clues about its reasoning process by checking the program and inspecting the output of its modules.

In this paper we apply the idea of NMN to video question answering and propose \textbf{STAIR}, a \textbf{S}patial-\textbf{T}emporal Reasoning model with \textbf{A}uditable \textbf{I}ntermediate \textbf{R}esults.

We define a set of basic video-text sub-tasks for video QA, such as localizing the time span of actions in the video, recognizing objects in a video clip, etc. We use a sequence-to-sequence program generator to decompose a question into its reasoning process, which is a hierarchical combination of several sub-tasks, and formalize this reasoning process into a formal-language program. Note that though the program generator requires question-program pairs to train, in practice we found that the program generator trained on AGQA2 \cite{GrundeMcLaughlin2022AGQA2} question-program pairs (which is publicly available) can generate plausible programs for questions from other datasets, so no further manual efforts are required to apply STAIR on video QA datasets without program annotations.

We also design a set of lightweight neural modules to complete each of these sub-tasks. These neural modules can be dynamically assembled into a neural module network according to the program. Then the neural module network takes video feature and text feature from a video encoder and a text encoder as input, and outputs a representation of the question after reasoning, which is then used by a classifier to generate the final answer. Different from most prior works of neural module networks, our neural modules return intermediate results specific to their intentions instead of always returning attention maps. Here we use the term ``auditable'' to describe that we can get the exact answer of each sub-task with no further actions required, which greatly increases the explainability of our method, and these intermediate results can also serve as prompts to improve the accuracy of pre-trained models.
We also introduce intermediate supervision to make the intermediate results more accurate by training neural modules with ground truth intermediate results.

We conduct experiments on the AGQA dataset \cite{GrundeMcLaughlin2021AGQA, GrundeMcLaughlin2022AGQA2}, a large-scale, real-world video question answering dataset with most questions of it require combinational temporal and logical reasoning to answer, for a detailed analysis of STAIR.
We also conduct experiments on STAR \cite{Wu2021STAR} and MSRVTT-QA \cite{Gao2018MotionAppearanceCN} to test the feasibility of STAIR on datasets without human annotations of programs. In summary, the contributions of this paper include:

\begin{itemize}
    \item We propose STAIR, a video question answering model based on neural module networks, which excels at solving questions that require combinational temporal and logical reasoning and is highly interpretable. We define sub-tasks for video QA, and design neural modules for the sub-tasks.
    \item We introduce intermediate supervision to make the intermediate results of the neural modules more accurate.
    \item We conduct extensive experiments on several video question answering tasks to demonstrate its performance, explainability, possibility to collaborate with pre-trained models, and applicability when program annotations are not available.
\end{itemize}

\section{Related Works}

\paragraph{Video Question Answering.}
Recent advances in video question answering methods can be roughly divided into four categories:
(1) \textbf{Attention based} methods \cite{Zhang2019OpenEndedLV,Li2019BeyondRP,Kumar2019LeveragingTA} that adopt spatial and/or temporal attention to fuse information from question and video;
(2) \textbf{Memory network based} methods \cite{Xu2017VideoQA,Gao2018MotionAppearanceCN,Fan2019HeterogeneousME,Kim2019ProgressiveAM} that use recurrent read and write operations to process video and question features;
(3) \textbf{Graph based} methods \cite{Jin2021AdaptiveSG,seo2021attend,Xiao2021VideoAC, Cherian2022251D, Park2021BridgeTA,Zhao2022Collaborative} that process videos as (usually object level) graphs and use graph neural networks to obtain informative video representations; and 
(4) \textbf{Pre-trained models} \cite{Lei2021LessIM,Fu2021VIOLETE,Zellers2021MERLOTMN,Zellers2022MERLOTRN,Wang2023Vstar} that pre-train a model in self-supervised manner with a mass of video-text multi-modal data. Recently, many works also try to solve video QA in zero-shot settings using large pre-trained transformer-based models \cite{Alayrac2022FlamingoAV, Li2023VideoChatCV, Zhang2023VideoLLaMAAI, Lyu2023MacawLLMML}. Though many works have reported good video understanding and response generation abilities of their models, these models require massive computing resources to pre-train, and their training videos/questions are relatively simple in terms of temporal reasoning, which means that these models are not robust at understanding and reasoning temporal information of videos.

Since there is usually redundant information in the video, Some works \cite{Kim2020ModalitySA, Gao2022MISTMI, Li2022InvariantGF} also study helping the model focus on key information by selecting video clips relevant to the question.

Though the above-mentioned methods have achieved outstanding performance, for most of these methods their performance tends to drop when evaluating on questions that require complicated logical reasoning or counterfactual questions and are difficult to interpret. To tackle these problems, some works use neural symbolic approach \cite{yi2019clevrer} \cite{Qian2022DynamicSM} or construct physics models \cite{Ding2021DynamicVR,chen2021grounding}.

\paragraph{Neural Module Networks.} 
Neural Module Networks (NMN) have been widely used in image question answering \cite{Andreas2015NeuralMN,Hu2017LearningTR,Johnson2017InferringAE,Mascharka2018TransparencyBD,Hu2018ExplainableNC}. These methods explicitly break down questions into several sub-tasks and solve each of them with a specifically-designed neural network (module). Attention maps or image representations are used to pass information among modules. Neural Module Networks are generally more interpretable, and excel at tasks that require compositional spatial reasoning such as SHAPES \cite{Andreas2015NeuralMN} and CLEVR \cite{Johnson2016CLEVRAD}.
A more advanced NMN for image-text tasks is the recently-proposed Visual Programming \cite{Gupta2023VisualProgramming}.
Taking advantage of several off-the-shelf models such as CLIP \cite{Radford2021LearningTV}, GPT-3 \cite{Brown2020LanguageMA} and Stable Diffusion \cite{rombach2021highresolution}, Visual Programming is capable of performing image QA, object tagging, and natural language image editing without further training.

Contrary to the intense research efforts of NMNs on image QA, there are significantly fewer works that focus on video QA \cite{Le2022VGNMNVN, Qian2022DynamicSM}. Though sharing the same motivation, it is non-trivial to define the sub-tasks and design their corresponding modules for video modality, which is one of the main contributions of our work. The work most similar to ours is DSTN \cite{Qian2022DynamicSM}, which also uses neural module network for video QA. But our work is significantly different from theirs in better performance, better explainability, the usage of intermediate supervision, the ability to collaborate with pre-trained models, and verifying its applicability when program annotations are not available.

\section{Methodology}
\begin{figure*}[t]
    \centering
    \includegraphics[width=0.9\textwidth]{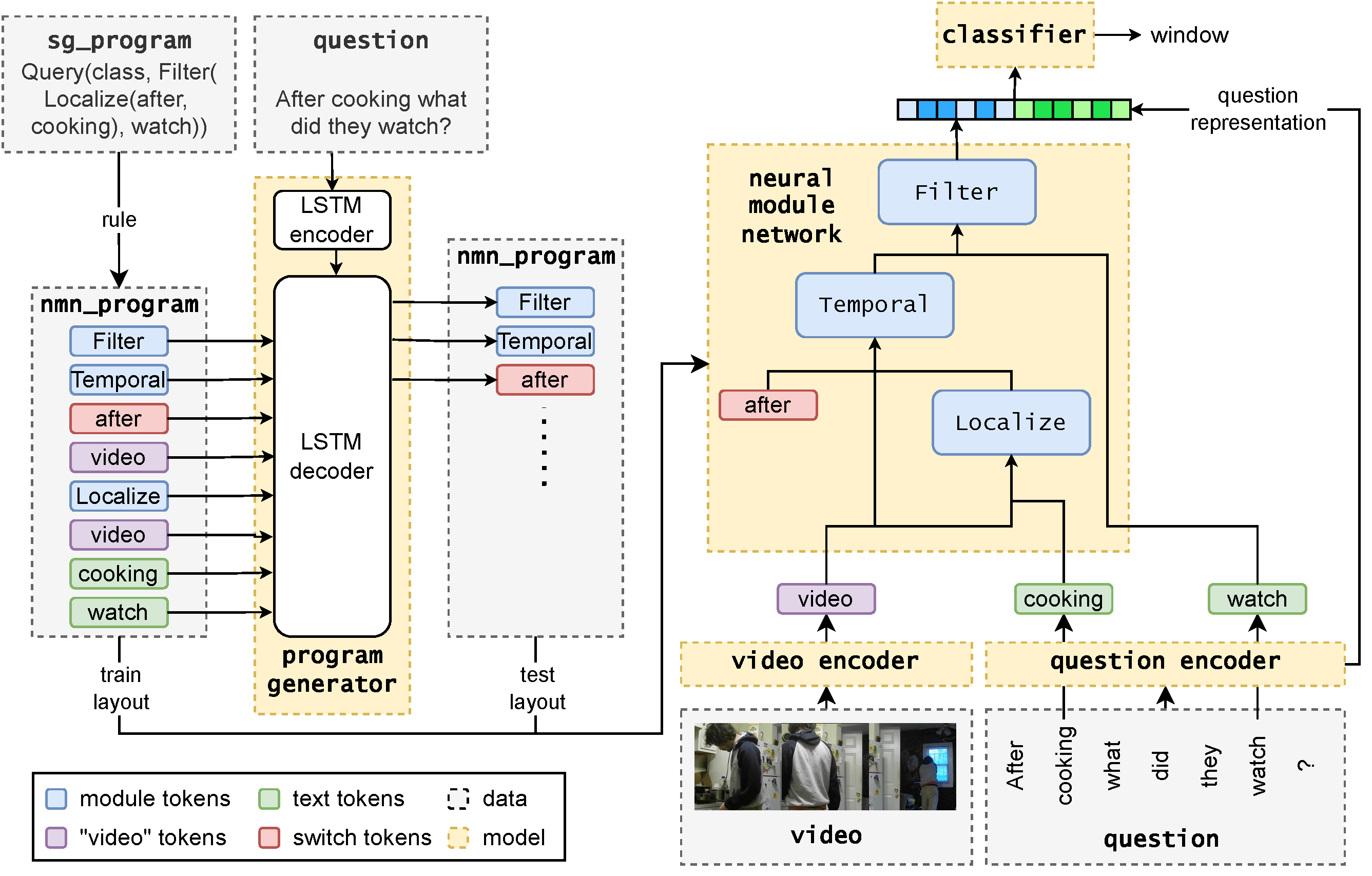}
    \caption{Overview of STAIR.}
    \label{fig:overview}
\end{figure*}

In this section, we describe the details of STAIR. STAIR takes as input a video feature $x_v \in \mathbb{R}^{T \times hid_V}$ with $T$ frames encoded by a pre-trained visual feature extractor and a question $x_q$ with $L$ words, and selects an answer $a$ from a fixed set of all possible answers $\mathcal{A}$. STAIR consists of the following components: 
(1) a bi-directional LSTM \textbf{video encoder} ${ENC}_{vid}$ which models the temporal relationship of the video feature and transforms it into the common hidden space $v = {ENC}_{vid}(x_v), v \in \mathbb{R}^{T \times H}$; (2) a bi-directional LSTM \textbf{text encoder} ${ENC}_{txt}$ which extracts the sentence-level and token-level question feature as $(q, t)={ENC}_{txt}(x_q), q \in \mathbb{R}^H, t \in \mathbb{R}^{L \times H}$; (3) \textbf{a collection of neural modules} $\{f_m\}$, each of which has a set of associated parameters $\theta_m$, performs a specific sub-task, and can be combined into a neural module network; and (4) a two-layer \textbf{classifier} $\phi(\cdot)$ that predicts the final answer. Besides, a \textbf{program generator} $p = gen(x_q)$ is trained individually to predict the program that determines the layout of the modules given a question $x_q$. The overview of the model is shown in Figure \ref{fig:overview}.

\subsection{Neural Modules}
As mentioned above, our solving process of the questions can be decomposed into several sub-tasks. For example, to answer the question \textit{"After cooking some food what did they watch?"}, there are 3 sub-tasks to solve: first localize the clips among the entire video when the people are cooking, then navigate to clips that happen after the cooking clips, and finally focus on these clips to find out the object that the people are watching.

Our STAIR contains 16 neural modules implementing different sub-tasks. All of these modules are implemented by simple neural networks such as several linear layers or convolutional layers. Their inputs, outputs, and intended functions are very diverse, including \texttt{Filter} module that finds objects or actions from a given video clip, \texttt{Exists} module that determines whether an object exists in the results of \texttt{Filter}, \texttt{Localize} module that finds in which frames an action happens, to name a few. The intentions and implementation details of all modules are listed in the Appendix. Different from most of the previous works of neural module networks, the inputs and outputs of our modules are not always the same (e.g., attention maps on images/videos), but are determined by the intentions of each module. Take the module \texttt{Filter(video,objects)} as an example, it intends to find all objects that appear in the video. Instead of returning an attention map showing when the objects occur, in our implementation it must return a feature vector from which we can predict the names of all objects in the video. This design leads to significantly better explainability and reliability, as we can know the exact objects it returns by only inspecting the output.

\subsection{Programs and the Program Generator}
The design of the program is inspired by the AGQA dataset. In AGQA, each question is labeled with a program consisting of nested functions indicating the sub-tasks of this question, and each video is tagged with a video scene graph from Charades and Action Genome \cite{sigurdsson2016hollywood,Ji2019ActionGA}. The answer can be acquired by executing the program on the scene graph. We use a rule-based approach to convert the labeled program to a sequence of program tokens, which is the Reverse Polish Notation of the tree-structured layout of our modules. \footnote{For details of this rule-based approach please refer to our code.} To avoid confusion hereafter we refer to the program before and after conversion as \texttt{sg\_program} and \texttt{nmn\_program}. Note that though \texttt{nmn\_program} is designed according to \texttt{sg\_program} in AGQA, it also works on other video question answering tasks as shown in Section \ref{sec:other_task}.

Program tokens in \texttt{nmn\_program} can be categorized into 4 types: (1) \textbf{module tokens} which corresponds to a neural module, e.g., \texttt{Filter}, \texttt{Localize}; (2) \textbf{the ``video'' token} that represents the video feature $v$; (3) \textbf{text tokens} which corresponds to a text span in the question $x_q$, e.g., ``watch'', ``cooking some food''; and (4) \textbf{switch tokens} which are keywords switches between the branches in a module, e.g., ``max'', ``after'', ``fwd''(``forward'').

As \texttt{nmn\_program}s are not provided during inference, we need to train a program generator to learn the mappings from questions to \texttt{nmn\_program}s. We tried fine-tuning a FLAN-T5-large \cite{wei2021finetunedLM}, but this problem is easy as a simple bi-directional LSTM encoder-decoder model with attention can predict exactly the right \texttt{nmn\_program} for more than 98\% of the questions in AGQA, so we decide to use the light-weight LSTM here.

\begin{figure}[t]
    \centering
    \includegraphics[width=0.5\textwidth]{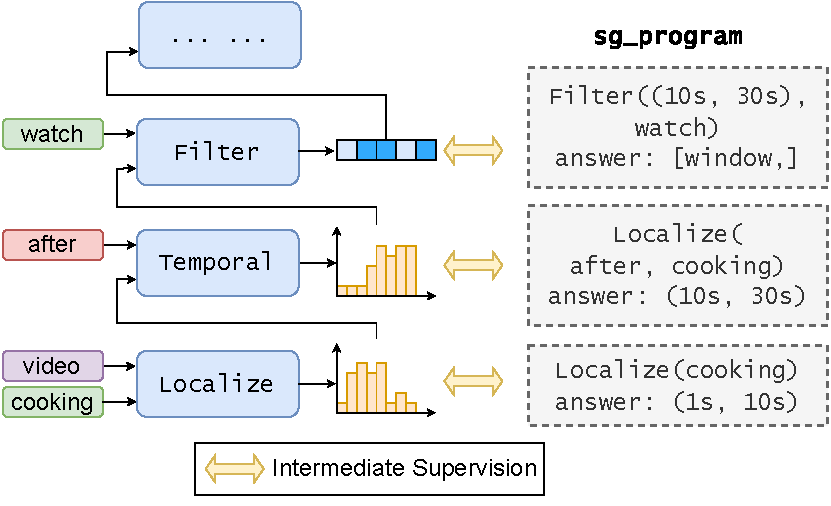}
    \caption{A Diagram of Intermediate Supervision.}
    \label{fig:intermediate_supervision}
\end{figure}

\subsection{Intermediate Supervision}
Previous works mentioned that sometimes modules in the neural module networks do not behave as we expected and thus can't provide meaningful intermediate outputs for us to understand its reasoning steps despite predicting the final answer correctly \cite{Hu2017LearningTR}. To mitigate this problem, we use \textbf{intermediate supervision} to induce supervision to intermediate modules. An example of intermediate supervision is shown in Figure \ref{fig:intermediate_supervision}.
Given that \texttt{nmn\_program} is obtained by converting \texttt{sg\_program} using a rule-based approach, we can record the correspondence between functions in \texttt{sg\_program} and modules in \texttt{nmn\_program}. Then we execute \texttt{sg\_program} on the video scene graph and take the return value of functions as ground truth answers of corresponding modules. \footnote{As the authors of AGQA and Action Genome do not release their code of acquiring answers via scene graphs, we have to implement these functions by ourselves. For about 5\% of all training examples, our implementation can't return the correct final answer given \texttt{sg\_program} and scene graph of the corresponding video, so we don't use intermediate supervision on them.} We use intermediate supervision for all but the first module in \texttt{nmn\_program} (i.e., the root module in the tree structure), as the first module is already directly supervised by the answer. Note that intermediate supervision does not always improve the model's performance, as its main purpose is to make the outputs of intermediate modules more accurate.
Depending on the data type, we use different criteria to calculate the intermediate supervision loss $\mathcal{L}^{IS}$ between the gold answer and module prediction, which is elaborated in the Appendix.

\subsection{Training Procedures}
The program generator is trained individually, and the main model, including video encoder, text encoder, neural modules, and classifier are trained in an end-to-end manner. 

Generating \texttt{nmn\_program} is considered as a sequence-to-sequence task. A model $gen(\cdot)$ takes question $x_q$ as input and generate \texttt{nmn\_program} $\hat{p}$ in an auto-regressive manner:

\begin{equation}
logP(\hat{p}|x_q) = \sum_t log(\hat{p_t} | x_q, \hat{p}_{<t})
\end{equation}

and the loss $\mathcal{L}^{GEN}$ is calculated using the negative log likelihood of ground truth \texttt{nmn\_program} $p$:

\begin{equation}
\mathcal{L}^{GEN}=-\sum_t log(p_t| x_q, p_{<t})
\end{equation}

When training the main model, the ground truth \texttt{nmn\_program} of train and valid set, or the \texttt{nmn\_program} generated by the program generator of test set is used to assemble the neural modules ${f_m}$ into a tree-structured neural module network. The classifier loss $\mathcal{L}^{CLS}$ is calculated using the ground truth answer $a$ and the predicted logits $\hat{a}$ over all candidate answers produced by the classifier as:

\begin{equation}
\mathcal{L}^{CLS}=l_{ce}(\hat{a}, a)
\end{equation}

The total loss of the main model is $\mathcal{L}=\mathcal{L}^{CLS}+\eta \mathcal{L}^{IS}$, where $\eta$ is a hyper-parameter balancing the classifier loss and the intermediate supervision loss.

\section{Experiments}
We evaluate STAIR mainly on AGQA balanced dataset \cite{GrundeMcLaughlin2021AGQA}, as it is a large-scale, real-world video QA dataset with most questions in it requiring comprehensive temporal reasoning to answer. AGQA balanced dataset contains 3.9M question-answer pairs with 9.6K videos. Each question is associated with a program that describes the reasoning steps to answer the questions. Videos in AGQA are from Charades \cite{sigurdsson2016hollywood}, a diverse human action recognition dataset collected by hundreds of people in their own homes. Each video is annotated with a video scene graph containing spatial and temporal information about actions and objects from Action Genome \cite{Ji2019ActionGA}. AGQA is very challenging, as even state-of-the-art deep learning models perform much worse than humans. We also evaluate on AGQA2 balanced dataset \cite{GrundeMcLaughlin2022AGQA2} which contains 2.27M question-answer pairs selected with a stricter balancing procedure and is even more challenging than AGQA. Following \cite{Le2020HierarchicalCR}, we leave 10\% of the train set out as valid set, and require videos in train/valid set to be different.

\subsection{Model Implementations}
\paragraph{Implementation.} We used two different video features in our experiments. One is the standard video features provided by the AGQA dataset, including appearance features $x_v^a \in \mathbb{R}^{8 \times 16 \times 2048}$ extracted from ResNet-101 pool5 layer\cite{He2015DeepRL}, and motion features $x_v^m \in \mathbb{R}^{8 \times 2048}$ extracted from ResNeXt-101\cite{Xie2016AggregatedRT}. We use mean pooling on the second dimension of $x_v^a$ and concatenate it with $x_v^m$ to obtain the final video feature $x_v \in \mathbb{R}^{8 \times 4096}$. We name this video feature ``RX''. However, as the official RX feature only has 8 frames on temporal dimension which is insufficient for complicated temporal reasoning, we also extract a video feature ourselves. We sample frames from videos with a frame rate of 24 fps, and use an I3D model pre-trained on Kinetics \cite{Carreira2017QuoVA} to extract a 1024-d feature for every consecutive 16 frames. We clip the temporal dimension length to 64, so the final video feature is $x_v \in \mathbb{R}^{T \times 1024}, T\leq64$. We name this video feature ``I3D''.

STAIR is trained with batch size 32, initial learning rate 2e-4 and decays linearly to 2e-5 in 200k steps. $\eta$ is set as 1. STAIR is trained on a Nvidia A100 GPU, and it takes about 2 epochs (30 hours) on average for a single run.

\paragraph{Baselines.} We compare \textbf{STAIR} with and without intermediate supervision (\textbf{-IS}) with several baselines. We compare with 3 representative video QA models: \textbf{HME} \cite{Fan2019HeterogeneousME} is a memory-network-based model to encode video and text features; \textbf{HCRN} \cite{Le2020HierarchicalCR} uses conditional relational networks to build a hierarchical structure that learns video representation on both clip level and video level; \textbf{PSAC} \cite{Li2019BeyondRP} uses both video and question positional self-attention instead of RNNs to model dependencies of questions and temporal relationships of videos.
To compare with models that explicitly model the multi-step reasoning process, we also compare with \textbf{DSTN} \cite{Qian2022DynamicSM}, a neural module   network concurrent to our work, and \textbf{MAC} \cite{hudson2018compositional} which performs iterative attention-based reasoning with a recurrent ``Memory, Attention and Composition'' cell. We make minor modifications on the attention of MAC to attend to 2-D $(T \times dim_V)$ temporal features instead of 3-D $(H \times W \times dim_V)$ spatial features.

\begin{table}[t]
    \centering
    \setlength{\abovecaptionskip}{5pt}
    \resizebox{0.99\columnwidth}{!}{
    \begin{tabular}{cc|ccc|c}
            \hline
            \textbf{Methods} & \textbf{Video} & \textbf{Binary} & \textbf{Open} & \textbf{Overall} & \textbf{\#Prm} \\
            \hline
            PSAC \dag & RX & 53.56 & 32.19 & 42.44 & 39M \\ 
            HME \dag & RX & 57.21 & 36.57 & 46.47 & 42M \\ 
            HCRN \dag & RX & 56.01 & 40.27 & 47.82 & 41M \\ 
            MAC & RX & 57.74 & 41.24 & 49.15 & 16M \\
            DSTN-E2E \dag & RX & 57.38 & 42.43 & 49.60 & 36M \\
            STAIR & RX & 59.07 & \textbf{43.08} & 50.75 & 21M \\
            STAIR-IS & RX & \textbf{60.15} & 42.84 & \textbf{51.14} & 21M \\
            \hline
            MAC & I3D & 58.19 & 46.84 & 52.28 & 10M \\
            STAIR & I3D & 60.18 & 47.24 & 53.45 & 14M \\
            STAIR-IS & I3D & \textbf{62.37} & \textbf{48.32} & \textbf{55.06} & 15M \\
            \hline
    \end{tabular}}
    \caption{Results of AGQA. \dag: Results from \cite{Qian2022DynamicSM}. \#Prm denotes number of parameters. \#Prm of MAC varies slightly with its number of steps, here we show \#Prm of a 12-step model.}
    \label{tab:agqa1}
\end{table}

\begin{table}[t]
    \centering
    \setlength{\abovecaptionskip}{5pt}
    \setlength{\tabcolsep}{4pt}
    \begin{tabular}{cc|ccc}
    \hline
    \textbf{Methods} & \textbf{Video} & \textbf{Binary} & \textbf{Open} & \textbf{Overall} \\
    \hline
    MAC & I3D & 54.72 & 44.96 & 49.67 \\
    STAIR & I3D & \textbf{57.13} & \textbf{47.07} & \textbf{52.06} \\
    STAIR-IS & I3D & 56.48 & 46.41 & 51.41 \\
    \hline
    \end{tabular}
    \caption{Results of AGQA2.}
    \label{tab:agqa2}
\end{table}

\begin{table}[t]
    \centering
    \setlength{\abovecaptionskip}{5pt}
    \setlength{\tabcolsep}{4pt}
    \begin{tabular}{c|ccc}
    \hline
    \multirow{2}{*}{\textbf{Methods}} & \textbf{Filter} & \textbf{Localize} & \textbf{Temporal} \\
     & (R@1/5) & (IoU) & (IoU) \\
    \hline
    Baseline & 0.11/0.43 & 0.16 & 0.13 \\
    STAIR & 0.12/0.30 & 0.19 & 0.35 \\
    STAIR-IS & \textbf{0.25/0.50} & \textbf{0.23} & \textbf{0.40} \\
    \hline
    \end{tabular}
    \caption{Performances of Filter, Localize and Temopral modules.}
    \label{tab:interpret}
\end{table}

\subsection{Model Performance}
Table \ref{tab:agqa1} shows the accuracy of all models on binary, open-ended and all questions of AGQA. STAIR outperforms all other baselines when using the same video feature, demonstrating the effectiveness of our approach. All models using the I3D video feature outperform their counterparts that use the RX feature, which shows the higher quality of I3D features. We also find that intermediate supervision does not always improve the performance of STAIR, probably due to the coordination problems among the losses of multi-task learning. However intermediate supervision does improve the model's explainability by making the output of intermediate results more accurate, which is shown in the next subsection. We also compare STAIR with the strongest baseline MAC using the I3D video feature on AGQA2, and the results are shown in Table \ref{tab:agqa2}.

\subsection{Evaluation and Visualization of Modules' Intermediate Output}
As our STAIR is based on neural module networks, it enjoys good interpretability while performing well. 
To demonstrate the interpretability of STAIR, we evaluate the intermediate results of \texttt{Filter}, \texttt{Localize} and \texttt{Temporal} modules, as these modules occurs at high frequency, and the outputs of them are intuitive and easy to inspect.

\texttt{Filter} module is designed to find objects and actions in the video or related to a given verb. To check the correctness of the output from \texttt{Filter} module, we use Recall@$N$ in a retrieval task as the evaluation metric. We calculate a candidate representation for each of the 214 candidate answers, and use cosine similarity between the output of \texttt{Filter} module and candidate representations to select a list of $N$ most likely predictions. If one of the predicted items occurs in the list of ground truth action(s)/object(s), we count it as a successful retrieval. We use the most frequently occurring $N$ actions/objects as baseline results.

\texttt{Localize} module is designed to find when an action happens in a video. We use $IoU_{att}$ as the evaluation metric. Given the predicted and ground truth attention scores $att_{p}, att_{g} \in \mathbb{R}^T$, the metric $IoU_att$ is calculated as $IoU_{att}=sum(min(att_{p}, att_{g})) / sum(max(att_{p}, att_{g}))$, where $max$ and $min$ are element-wise operations. We use uniform distribution as baseline attention scores: $att_{b} \sim \mathbf{U}(0,1) \times T$.

\texttt{Temporal} module is designed to transform the attention scores according to the switch keyword $s$. We use the same metric $IoU_{att}$ to evaluate the attention scores output $att_{out}$. Inspired by \cite{Qian2022DynamicSM}, we randomly sample two frames as the start and end frames as baseline results. Specially, the start frame is always the first frame when $s=$ `before', and the end frame is always the last frame when $s=$ `after'.

Table \ref{tab:interpret} shows the results. STAIR performs baseline on most metrics except R@5 of \texttt{Filter} module, which indicates that STAIR is capable of providing meaningful intermediate results, and training with intermediate supervision can make the intermediate results more accurate.

We also visualize the reasoning process of STAIR on some real examples in the test set in the Appendix.

\begin{table*}[t]
    \setlength{\abovecaptionskip}{5pt}
    \centering
    \begin{tabular}{c|ccc|ccc|c}
        \hline
        \multirow{2}{*}{\textbf{Methods}} & \multicolumn{3}{c|}{\textit{AGQA}} & \multicolumn{3}{c|}{\textit{AGQA2}} & \multirow{2}{*}{\textbf{\#Params}} \\
        & \textbf{Binary} & \textbf{Open} & \textbf{Overall} & \textbf{Binary} & \textbf{Open} & \textbf{Overall} & \\
        \hline
        STAIR & 60.18 & 47.24 & 53.45 & 57.13 & 47.07 & 52.06 & 14.97M \\
        STAIR-IS & 62.37 & 48.32 & 55.06 & 56.48 & 46.41 & 51.41 & 15.11M \\
        GPT-2 & 63.94 & 50.88 & 57.14 & 58.10 & 47.90 & 52.96 & 127M \\
        Violet & 60.87 & \textbf{52.88} & 56.72 & 50.28 & \textbf{49.93} & 50.11 & 160M \\
        \makecell[c]{GPT-2+ \\ STAIR-IS} & \textbf{64.26} & 50.97 & \textbf{57.34} & \textbf{60.46} & 49.86 & \textbf{55.13} & \makecell[c]{127M+ \\ 15.11M} \\
        \hline
    \end{tabular}
    \caption{Results of AGQA and AGQA2, comparing with pre-trained models.}
    \label{tab:pretrain}
\end{table*}

\begin{table}[t]
    \centering
    \begin{tabular}{c|c}
        \hline
        \textbf{Methods} & \textbf{Overall} \\
        \hline
        Video-ChatGPT & 35.09 (0.76) \\
        + STAIR-IS & \textbf{40.43} (0.89) \\
        \hline
    \end{tabular}
    \caption{Results of Video-ChatGPT on AGQA2.}
    \label{tab:zs_agqa2}
\end{table}

\subsection{Compatibility with Pre-trained Models}
Pre-trained models, including text-only ones and multi-modal ones, have achieved state-of-the-art performance on many question answering tasks. Here we first compare STAIR with a single-modal pre-trained model \textbf{GPT-2} \cite{Radford2019LanguageMA}, and a video-text pre-trained model \textbf{Violet} \cite{Fu2021VIOLETE}.
For GPT-2, we prepend I3D video features to questions and assign different token type embeddings following \cite{Li2020BridgingTA}.
For Violet, we sampled $T=10$ video frames, resize them into $224 \times 224$, and split them into patches with $W \times H = 32 \times 32$. Though we can't use the pre-trained temporal position embedding as our $T=10$ is larger than $T=4$ in the pre-training stage and $T=5$ for downstream tasks in the original paper, we find that this gives better results. Table \ref{tab:pretrain} shows that on AGQA STAIR still underperforms GPT-2 and Violet, probably due to significantly fewer parameters and the absence of pre-training.
However, the performance gap between STAIR and the pre-trained models on AGQA2 is smaller, probably due to the language bias being further reduced and it's harder for pre-trained models to find textual clues to solve the questions. 

To combine STAIR with pre-trained models, we use a straightforward method: we modify the questions to add the intermediate results of our neural modules to the input of pre-trained models as prompts. We get the top 1 candidate result for every \texttt{Filter} module in STAIR-IS using methods described in intermediate output subsection, and concatenate it with its keyword inputs. As \texttt{Filter} modules with lower levels have higher accuracy, we sort all \texttt{Filter} modules in ascending order of level and take only the first $\mathbf{P}$ modules into account, where $\mathbf{P}$ is selected in \{1,3,5\} by valid set performance. Take the following question as an example: \textit{What did they take while sitting in the thing they went above?}. To answer this question, the corresponding \texttt{nmn\_program} contains one \texttt{Filter} module with parameter $(video, above)$ and returns the result ``bag". So the modified question becomes: \textit{above bag. What did they take while sitting in the thing they went above?}. This can reduce the difficulty of questions by providing answers to some sub-tasks so it requires fewer steps to answer them.
We use this method on the best-performing GPT-2 and denote it as \textbf{GPT-2+STAIR-IS}. Experiments show that with the help of these intermediate outputs, the performance of GPT-2 is further improved. It is also an evidence of the usefulness of the intermediate results.

Given the recent rapid development of multi-modal large pre-trained models, we also report the results of zero-shot \textbf{Video-ChatGPT} \cite{Maaz2023VideoChatGPTTD}, a video-text pre-trained model which is claimed to be optimized for temporal understanding in videos, and \textbf{Video-ChatGPT + STAIR-IS} in Table \ref{tab:zs_agqa2}. Following \cite{Maaz2023VideoChatGPTTD}, Video-ChatGPT is not fine-tuned, and we benchmark its performance on AGQA2 with the evaluation pipeline using GPT-3.5. As it is unfeasible to test on the entire test set of AGQA2 with 660K questions, we randomly sample 1\% (6.6K questions), repeat the experiment for 3 times, and report the average accuracy and standard deviation.

\begin{table}[t]
\setlength{\abovecaptionskip}{5pt}
\setlength{\tabcolsep}{3pt}
    \centering
    \begin{tabular}{c|cccc}
        \hline
        \textbf{Methods} & \textbf{Int.} & \textbf{Seq.} & \textbf{Pre.} & \textbf{Fea.} \\
        \hline
        CNN-BERT \dag & 33.59 & 37.16 & 30.95 & 30.84 \\
        L-GCN \dag & 39.01 & 37.97 & 28.81 & 26.98 \\
        HCRN \dag & \textbf{39.10} & 38.17 & 28.75 & 27.27 \\
        STAIR & 33.20 & \textbf{39.16} & \textbf{38.41} & \textbf{31.30} \\
        \hline
        ClipBERT \dag & 39.81 & 43.59 & 32.34 & 31.42 \\
        \hline
    \end{tabular} 
    \caption{Accuracy on STAR test set, categorized by question type. \dag: Results from \cite{Wu2021STAR}}
    \label{tab:star}
\end{table}

\begin{table}[t]
    \centering
    \begin{tabular}{c|c}
        \hline
        \textbf{Methods} & \textbf{MSRVTT-QA} \\
        \hline
        Co-Memory \cite{Gao2018MotionAppearanceCN} & 32.0 \\
        HME \cite{Fan2019HeterogeneousME} & 33.0 \\
        HCRN \cite{Le2020HierarchicalCR} & \textbf{35.6} \\
        STAIR & 34.8 \\
        \hline
        ClipBERT \cite{Lei2021LessIM} & 37.4 \\
        \hline
    \end{tabular} 
    \caption{Accuracy on MSRVTT-QA test set.}
    \label{tab:msrvttqa}
\end{table}

\subsection{Experiments on Tasks Without Program Annotations} \label{sec:other_task}

One may question that the need for program annotations limits the usage of STAIR. However, this question can be resolved by verifying that program generators trained on AGQA can be used to generate programs for questions from other video QA datasets: since the program annotations of AGQA is already publicly available, no more manual efforts are required to apply STAIR on datasets without program annotations.

To resolve this question, we conduct experiments on STAR \cite{Wu2021STAR} and MSRVTT-QA \cite{Xu2017VideoQA}. We changed the program generator from an LSTM to a FLAN-T5-large \cite{wei2021finetunedLM} fine-tuned on AGQA2 question-\texttt{nmn\_program} pairs to make the program generator more generalizable. Please refer to the Appendix for details of the experiments. Surprisingly, though the program generator has never seen questions from STAR and MSRVTT-QA during training phase, it can generate executable programs for more than 95\% of the questions. Results are shown in Table \ref{tab:star} and Table \ref{tab:msrvttqa}. Though STAIR do not perform well on Interaction type of questions as they are too simple to take advantage of the compositional ability of the neural modules, it outperformes several video question answering baselines on Sequence, Prediction and Feasibility types of questions which requires spatial and temporal reasoning. Results on MSRVTT-QA shows that STAIR is also applicable to noisy, automatically-generated questions \cite{Lin2022TowardsFA}. However, it performs worse than the pre-trained ClipBERT and is only comparable with other simpler methods, as STAIR is designed for complex spatial-temporal reasoning while questions in MSRVTT-QA are mostly simple factoid questions. 

\section{Conclusion}
In this paper, we propose STAIR for explainable compositional video question answering. We conduct extensive experiments to demonstrate the performance, explainability, and applicability when program annotations are not available. Moreover, STAIR is more auditable compared with previous works, it returns direct, human-understandable intermediate results for almost every reasoning step, and can be used as prompts to improve the performance of pre-trained models. We also propose intermediate supervision to improve the accuracy of intermediate results.

Possible future directions include: training program generators without direct supervision of ground truth programs (e.g., with reinforcement learning like \cite{Mao2019TheNC}), better functional and structural designs of the neural modules (e.g., using more powerful pre-trained models), and applying on more video-text tasks other than QA.

\section{Acknowledgments}
This work was supported by National R\&D project of China under contract No. 2022YFC3301900.



\bibliography{aaai24}

\section{Appendix}

\subsection{Objectives of Intermediate Supervision}
For module $m$ of \texttt{Filter}, \texttt{ToAction} and \texttt{Superlative} types, as these modules return a vector $t$, and their corresponding functions return a set of ground truth action(s)/object(s) $Y=\{y\}$ (for \texttt{ToAction} and \texttt{Superlative} the size of this set is 1, and for \texttt{Filter} the size can be greater than 1). Our goal is to predict the ground truth set $Y$ using vector $t$ as a multi-label classification task. We borrow the idea from non-parametric classification \cite{Wu2018UnsupervisedFL}: we calculate a class prototype vector $v$ for each object/action $e$ of all $n$ objects/actions that appear in the ground truth sets in the mini-batch using the text encoder: $v_i={ENC}_{txt}(e_i), i=1 \dots n$. Then we calculate the probability $P(i|t)$ of $t$ is (or contains) action/object $e_i$ as 
\begin{equation}
P(i|t)=\frac{exp(t^Tv_i)}{\Sigma_{j=1}^nexp(t^Tv_j)}
\end{equation}

and use binary cross entropy to calculate the loss between the probability and the ground truth list for each module:
\begin{equation}
\mathcal{L}^{IS}_m = l_{bce}(P(\cdot|t),Y)
\end{equation}

For module $m$ of \texttt{ExistFrame}, \texttt{Localize} and \texttt{Temporal} types, as these modules return an attention vector, we consider the attention scores as the probability of a frame being ``selected'', and use binary cross entropy as the loss function. Formally, if the predicted attention vector is $att$ where $att_i \in [0,1]$ and the ground truth attention vector is $y$ where $y_i = \mathds{1}[ i \in [start, end) ]$ for $i = 0 \dots T-1$, we have

\begin{equation}
\mathcal{L}^{IS}_m = l_{bce}(att, y)
\end{equation}

For module $m$ of \texttt{Exists} and \texttt{Equals} types, as the return value of their corresponding function is binary  (yes/no), we additionally append a linear classification layer with parameters $W \in \mathbb{R}^{H \times 2}, b \in \mathbb{R}^2$ after the model output and use cross entropy loss. Formally, if the predicted representation is $t \in \mathbb{R}^H$ and the ground truth is a label $y \in \{0, 1\}$, we have

\begin{equation}
\mathcal{L}_m^{IS} = l_{ce}(Wt+b, y)
\end{equation}

Finally, the overall intermediate supervision loss is calculated as follows:

\begin{equation}
\mathcal{L}^{IS}=\sum_{m \in \texttt{nmn\_program[1:]}} \mathcal{L}_m^{IS}.
\end{equation}

\subsection{Details of Experiments on STAR and MSRVTT-QA}
When generating \texttt{nmn\_program}s, we use beam search with beam size 5 to generate 5 candidate programs for each question, sort them by probability in descending order, and choose the first executable candidate as \texttt{nmn\_program}. We discard questions with no available executable programs in train and valid set, and use a random feature as neural module network output in the test set. We found that about 15\% of text tokens in \texttt{nmn\_program} cannot be easily mapped to words in the question using simple rules, so we use the representation of all words in the question as the representation of this token.

For the multiple-choice questions in STAR, we use another LSTM with the same model structure as the question encoder as the choice encoder, and use the cosine similarity between choice and [nmn\_output; question\_representation] as the logits of the choices.

We use video features extracted by the same appearence encoder as HCRN \cite{Le2020HierarchicalCR} (ResNet-101 \cite{He2015DeepRL}) for a fair comparison with baselines.

\medskip

\begin{table*}[t]
\centering
\resizebox{\textwidth}{!}{
    \begin{tabular}{cm{4cm}cclc}
        \hline
        \textbf{Module} & \makecell[c]{\textbf{Intention}} & \textbf{Input} & \textbf{Output} & \makecell[c]{\textbf{Implementation}} & \textbf{Freq.} \\
        \hline
        And & performs AND on 2 representations & $t_1, t_2$ & $t_{out}$ & $t_{out}=min(t_1, t_2)$ & 0.02 \\
        \hline
        
        AttnVideo & returns a video representation weighted by an attention score &  $v_{in}, a_{in}$ & $v_{out}$ & $v_{out}=v_{in} \odot a_{in}$ & 1.19 \\
        \hline
        
        Choose & chooses a key representation that is more similar to the query & $t_q, t_{k1}, t_{k2}$ & $t_{ans}$ & 
            $ t_{ans}=\begin{cases}
                t_{k1} & cos(t_q, t_{k1}) > cos(t_q, t_{k2}) \\
                t_{k2} & \textit{otherwise} \\
                \end{cases}
            $ & 0.15 \\
        \hline
        
        Compare & returns the statement more likely to be true between 2 statements returned by Exists & $t_1, t_2$ & $t_{out}$ & $t_{out}=W[t_1;t_2]+b $ & 0.14 \\
        \hline
        
        Equals & judges whether two representations are semantically equal & $t_1, t_2$ & $t_{out}$ & $t_{out}=W[t_1;t_2]+b $ & 0.06 \\
        \hline
        
        Exists & judges whether an object exists in the feature returned by Filter & $t_q, t_k$ & $t_{out}$ & $t_{out}=W_2(W_1[t_q;t_k;t_q \odot t_k]+b_1)+b_2$ & 0.46 \\
        \hline
        
        ExistsFrame & returns the probability of an object exists frame-by-frame in the feature returned by FilterFrame & $t_{in}, v_{in}$ & $a_{out}$ & $a_{out}=(cos(t_{in}, v_{in})+1) \times 0.5$ & 1.29 \\
        \hline
        
        \makecell[c]{Filter \\ $[$objects | actions$]$} & finds all objects/actions in the video & \makecell[c]{$v_{in}$, \\ $s \in \{$\textit{obj, act}$\}$} & $t_{out}$ & $t_{out}=W_3sum(W_{s,2}(W_{s,1}v_{in}+b_{s,1})+b_{s,2},1)+b_3$ & 0.23 \\
        \hline
        
        \makecell[c]{Filter \\ $[$verb$]$} & finds all objects related to a given verb in the video & $t_{in}, v_{in}$ & $t_{out}$ & 
            \makecell[l]{
                $v'=W_2(W_1v_{in}+b_1)+b_2$ \\
                $v_{agg}=W_3[v';t_{in}]+b_3 \odot v_{in}$ \\
                $t_{out}=W_4sum(v_{agg},0)+b_4$
            } & 1.96 \\
        \hline
        
        \makecell[c]{FilterFrame \\ $[$objects | actions$]$} & finds all objects/actions in each frame & \makecell[c]{$v_{in}$, \\ $s \in \{$\textit{obj, act}$\}$} & $v_{out}$ & $v_{out}=W_3(W_{s,2}(W_{s,1}v_{in}+b_{s,1})+b_{s,2})+b_3$ & 0.01 \\
        \hline
        
        \makecell[c]{FilterFrame \\ $[$verb$]$} & finds all objects related to a given verb in each frame & $t_{in}, v_{in}$ & $v_{out}$ &
            \makecell[l]{
                $v'=W_2(W_1v_{in}+b_1)+b_2$ \\
                $v_{agg}=W_3[v';t_{in}]+b_3 \odot v_{in}$ \\ 
                $v_{out}=W_4v_{agg}+b_4$
            } & 1.29 \\
        \hline
        
        Localize & finds when an action happens in the video & $t_{in}, v_{in}$ & $a_{out}$ &
            \makecell[l]{
                $v'=W_2(W_1v_{in}+b_1)+b_2$ \\ 
                $t'=W_3t_{in}+b_3$ \\
                $a_{out} = (cos(v',t') + 1) \times 0.5 $
            } & 0.98 \\
        \hline

        Relate & returns the first/last video span & \makecell[c]{$a_{in}$, \\ $s \in \{$\textit{fwd, bkwd}$\}$} & $a_{out}$ & 
            $ a_{out}=\begin{cases}
                a_{in} + b & s=\textit{`fwd'} \\
                a_{in} - b & s=\textit{`bkwd'} \\
            \end{cases} $ & 1.19 \\
        \hline
        
        Superlative & finds the longer/shorter action of 2 given actions & \makecell[c]{$t_1, t_2, v_{in}$, \\ $s \in \{$\textit{max, min}$\}$} & $t_{out}$ & 
            \makecell[l]{
                $\{a'\}=\{$\textit{Localize}$(t_i, v_{in}), i=1,2\}$ \\
                $ w = \begin{cases}
                    \text{softmax}(\text{sum}(\{a'\})) & s=\mathit{max} \\
                    1-\text{softmax}(\text{sum}(\{a'\})) & s=\mathit{min} \\
                \end{cases} $ \\               
                $t_{out}=sum(w \odot \{t\}_{in},0)$ 
            } & 0.02 \\
        \hline
        
        Temporal & returns temporal weights of a video span according to the switch & \makecell[c]{$v_{in}, a_{in}$, \\ $s \in \{$\textit{before, after,} \\ \textit{while, between}$\}$} & $v_{out}, a_{out}$ & 
            \makecell[l]{
                $a_{out}=conv_{s}(a_{in})$ \\
                $v_{out}=W(a_{out} \odot v_{in})+b$ \\
            } & 0.98
        \\
        \hline
        
        ToAction & converts a verb and an object to an action & $t_{verb}, t_{noun}$ & $t_{out}$ & $t_{out}=W_2(W_1[t_{verb};t_{noun}]+b_1)+b_2$ & 0.72 \\
        \hline
        
        Xor & performs XOR on 2 representations & $t_1, t_2$ & $t_{out}$ & $t_{out}=W_1[t_1;t_2; abs(t_1-t_2)]+b_1$ & 0.02 \\
        \hline
        
        XorFrame & performs XOR on 2 attention maps & $a_1, a_2$ & $a_{out}$ & $a_{out}=abs(a_1-a_2)$ & 0.10 \\
        \hline
    \end{tabular}
}
    \caption{Intentions and implementation details of all modules. There are 4 types of variables: $v \in \mathbb{R}^{T \times H}$ denotes a video feature, which contains a feature vector of size $\mathbb{R}^H$ for each of the $T$ frames; $t \in \mathbb{R}^H$ denotes a text feature, which is usually an action/object or a set of actions/objects; $a \in \mathbb{R}^T$ denotes an attention map over the frames; and $s$ denotes a keyword that switches between the branches in a module. $sum(\cdot, n)$ denotes summing a tensor on dimension $n$, $conv(\cdot)$ denotes a 1-D convolutional network, and $\{\dotsc\}$ denotes a list of items. Activation functions and dropouts are omitted to avoid cluttering. Freq. denotes the average number of occurrences of the module in one program in train and valid set of AGQA.}
    \label{tab:modules}
\end{table*}

\begin{figure*}[t]
    \includegraphics[width=0.99\textwidth]{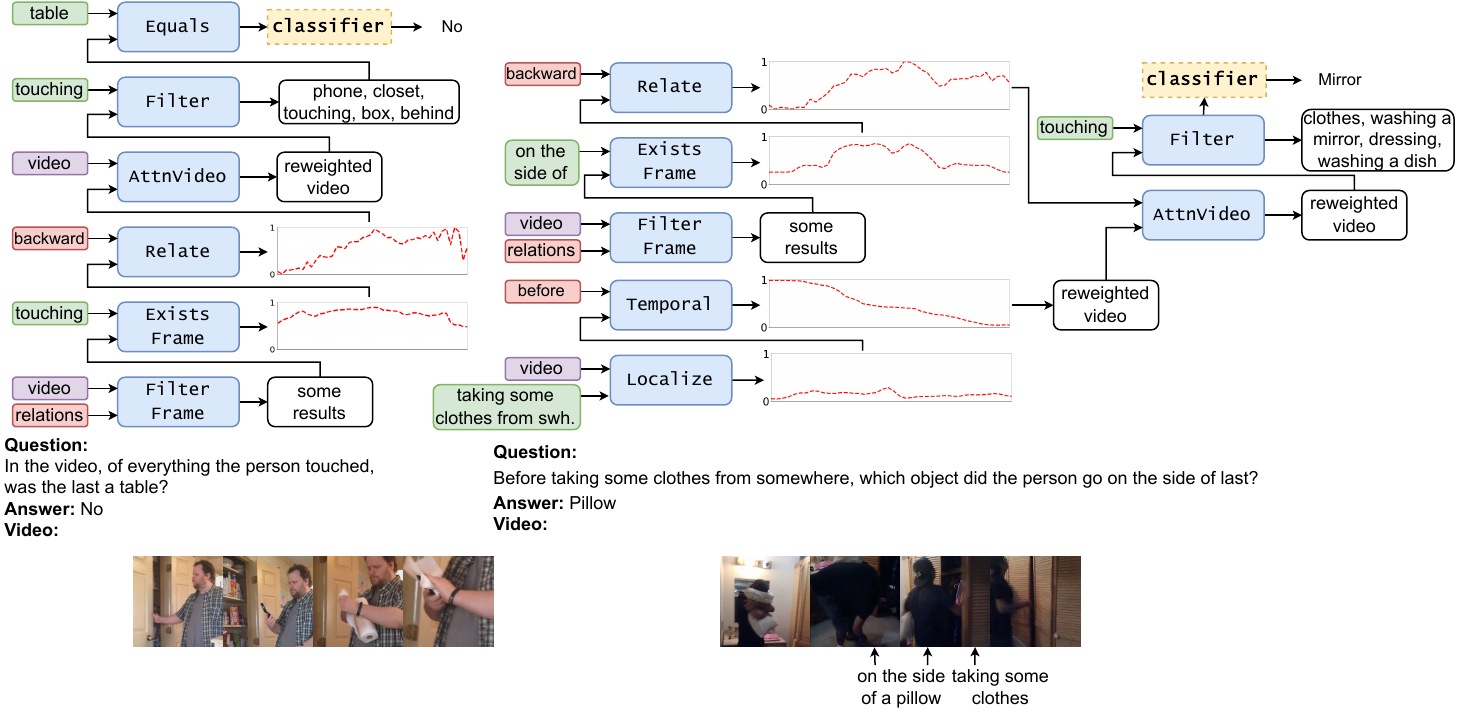}
    \caption{Examples of a successful case (left) and a failing case (right). In the first case, the person is touching things throughout the video, so the \texttt{ExistsFrame} module returns a uniform distribution on all the frames. The last 2 things the person touches are phone and tissue, though \texttt{Filter} module only finds one correct answer ``phone'', but as it is not equal to the choice ``table'', so \texttt{Equals} module returns the correct final answer ``No''. In the second case, \texttt{Localize} module successfully finds when the person is taking some clothes and \texttt{ExistsFrame} module successfully finds when the person is on the side of something, but \texttt{Filter} module fails to recognize the exact thing that is on the side of the person (probably due to low video quality and the pillow is blocked by the body). Outputs of \texttt{FilterFrame} modules are too complex to be visualized.}
    \label{fig:case}
\end{figure*}

\end{document}